\documentclass[]{ailab}

\usepackage{amsmath,amssymb}
\usepackage{graphicx}
\graphicspath{{figures/new_figures/}{figures/}{.}}
\usepackage{booktabs}
\usepackage{longtable}
\usepackage{array}
\usepackage{natbib}
\usepackage{xspace}
\usepackage{float} 
\usepackage{placeins} 
\setcitestyle{square,comma,numbers,sort&compress}

\usepackage{iftex}
\ifPDFTeX
  \DeclareUnicodeCharacter{2011}{\nobreakdash-}
\else
  \usepackage{newunicodechar}
  \newunicodechar{‑}{\nobreakdash-}
  \newunicodechar{’}{\textquoteright}
\fi

\providecommand{\ours}{\textsc{Long-Horizon-Terminal-Bench 2.0}\xspace}
\providecommand{\oursabbr}{\textsc{LHTB 2.0}\xspace}
\providecommand{\lhtbone}{\textsc{Long-Horizon-Terminal-Bench}\xspace}

\title{Long-Horizon-Terminal-Bench 2.0: Automating Diverse Task Construction and Measuring Sustained Agent Progress}

\author[1,2]{Zongxia Li\textsuperscript{$\dagger$}}
\author[1,3]{Zhongzhi Li\textsuperscript{$\dagger$}}
\author[1]{Yucheng Shi\textsuperscript{$\dagger$}}
\author[1,4]{Ruhan Wang}
\author[1,5]{Junyao Yang}
\author[6]{Chengsong Huang}
\author[2]{Zhichao Liu}
\author[2]{Xiyang Wu}
\author[7]{Anhao Li}
\author[4]{Yue Yu}
\author[8]{Ergan Shang}
\author[2]{Fuxiao Liu}
\author[2]{Tong Zheng}
\author[9]{Ninghao Liu}
\author[10]{Lichao Sun}
\author[1]{Haitao Mi}
\author[1]{LeoweiLiang}

\affiliation[1]{Tencent HY LLM Frontier}
\affiliation[2]{University of Maryland, College Park}
\affiliation[3]{University of Georgia}
\affiliation[4]{Indiana University}
\affiliation[5]{National University of Singapore}
\affiliation[6]{Washington University in St. Louis}
\affiliation[7]{University of Minnesota, Twin Cities}
\affiliation[8]{Carnegie Mellon University}
\affiliation[9]{The Hong Kong Polytechnic University}
\affiliation[10]{Lehigh University}

\email{zli12321@umd.edu, zongxiali@global.tencent.com, zl22754@uga.edu}

\contribution{$^\dagger$ Core contribution.}

\abstract{
Long-horizon terminal benchmarks measure whether an agent can sustain progress through extended workflows with process rewards, not only whether its final state passes a binary check. 
We present \ours{} (\oursabbr{}), an extension of \lhtbone that preserves its 46 original tasks while correcting task and evaluation-infrastructure defects and extending 31 tasks. 
\oursabbr{} contains 77 tasks across 13 domains and eight task types. Different from prior benchmark, the extended tasks are produced through a systematic reconstruction-task pipeline. 
We start from a working scientific or engineering artifact, then use a parameterized reference implementation to generate multiple scenarios, reveal six as behavioral evidence, hold out three for evaluation, and use dense progress ladders to reward partial progress while preventing memorization of public outputs.
We also extend the evaluation horizon from 90 minutes to three hours and record verifier checkpoints every 30 minutes, enabling direct measurement of sustained progress within a run. 
Across 1,145 scored model--task runs from 15 frontier models, GPT-6-astra leads with dense mean reward $0.69$ and 37 tasks at $R \geq 0.95$, followed by GPT-5.6-sol with dense mean reward $0.67$ and 32 tasks at the same threshold. 
On the 30 reconstruction tasks, binary verifier scores average $0.221$, whereas dense ladder scores average $0.653$, showing that outcome-only evaluation discards substantial intermediate progress. 
\oursabbr{} provides a broader, hardened, and process-aware testbed for studying reliable agent execution over hours.
}

\headercontent{{\sffamily\bfseries Project page:} \url{https://zli12321.github.io/LHTB/}}

\date{\today}

\begin{document}
\thispagestyle{firstheader}
\maketitle

\section{Introduction}
AI agents built on large language models are increasingly expected to operate beyond short, self-contained interactions: they must use tools, maintain state, recover from errors, and coordinate many dependent steps in workflows that last minutes or hours. 
This shift creates two related evaluation challenges. 
First, a final pass/fail outcome reveals whether an agent completed a task, but not whether it made sustained progress, stalled, or regressed over time. 
Second, as agent capabilities expand into new domains, manually constructing realistic and reliable long-horizon tasks becomes increasingly difficult to scale, and also increasingly difficult for humans to annotate, but easy to measure agents. 
Evaluating the next generation of agents therefore requires both scalable task construction and process-aware measurements that track progress throughout extended trajectories.

\lhtbone (LHTB) benchmark introduced 46 containerized tasks with dense, subtask-based grading for these workflows~\cite{li2026longhorizonterminalbenchtestinglimitsagents}. It established that agents can make meaningful partial progress even when they do not finish, but it also exposed several limitations. 
First, manually curating long-horizon tasks is expensive and difficult to scale across domains~\cite{chu2026terminalworldbenchmarkingagentsrealworld,song2026envscalerscalingtoolinteractiveenvironments}. 
Second, most existing benchmarks score only the final outcome of each run, hiding whether an agent makes sustained progress, plateaus, or regresses over time~\cite{li2026finalscoressystematicevaluation}.
Third, repeated evaluation revealed task-environment and verifier-integration defects that must be corrected before larger evaluations can be compared reliably.

We propose \ours{} (\oursabbr{}) to address the limitations of the previous benchmark. version. 
The benchmark retains the 46 original tasks, fixes task and harness defects, and adds 31 tasks, a total of 77-task suite spanning 13 domains and eight task types. 
The evaluation infrastructure now isolates checkpoint grading from the active agent, clears verifier artifacts between phases, audits successful trajectories for grader access, and retains only clean attempts. 
We also extend the experiment run to three-hour trials with verifier checkpoints every 30 minutes make it possible to measure how reward evolves within the same trajectory rather than comparing unrelated short and long runs.

The central technique for automating long-horizon and difficult task generation is through task reconstruction~\cite{li2026recursivesynthesislonghorizonterminal,li2026selfsupervisedscalingterminalenvironments}.
Starting from a known working, we identify parameter axes, execute a reference oracle over a scenario,  publish multiple input--output examples as behavioral evidence, and withhold multiple scenarios for grading. 
The agent receives the tools and evidence needed to infer the behavior, but not the reference implementation. 
%
This procedure contributes 30 new reconstruction tasks in science domains such as geospatial analysis, document processing, life sciences, chemistry and materials, physical simulation, signals and control, security, and related domains.

Our evaluation covers 1,145 scored model--task cells across 15 frontier models. On a dense all-77 basis, GPT-6-astra leads with mean reward $0.69$ and 37 tasks at $R\geq0.95$, followed by GPT-5.6-sol with mean reward $0.67$ and 32 tasks at the same threshold. GPT-6-astra's lead comes almost entirely from the reconstruction split; on the released workflows GPT-5.6-sol is slightly ahead. More importantly, the 30 reconstruction tasks average only $0.221$ under binary verifier scoring but $0.653$ under dense ladder scoring. The gap shows that binary evaluation frequently labels hours of useful progress as total failure. The checkpoints further show that $42\%$ of solves arrive after the first hour, that a one-hour budget would reorder the middle of the leaderboard, and that the fastest models often claim completion long before they finish.

The new benchmark extension includes:
\begin{itemize}
    \item \textbf{A broader and corrected long-horizon benchmark.} We extend \lhtbone from 46 to 77 tasks, covering 13 domains and eight task types while preserving the original suite.
    \item \textbf{Verifying the feasibility of a systematic task-construction pipeline.} We turn parameterized real-world artifacts into reconstruction tasks with public behavioral evidence, hidden scenarios, anti-memorization checks, and dense progress ladders.
    \item \textbf{Process-aware long-horizon progress evaluation.} We increase the budget from 90 minutes to three hours and record 30-minute checkpoints, enabling direct analysis of sustained progress, late completion, and regression.
    \item \textbf{A hardened evaluation protocol.} We fix task and harness bugs, isolate verifier execution, audit grader access, and replace affected results with clean reruns.
\end{itemize}

\section{From LHTB to \oursabbr{}}
\label{sec:background}

\subsection{Task formulation}

\ours{} follows the Terminal-Bench and Harbor formulation~\cite{merrill2026terminalbenchbenchmarkingagentshard}. Each task contains (1) a natural-language instruction, (2) a containerized terminal environment, (3) task and runtime configuration, and (4) a deterministic verifier or simulator. The instruction is the only specification shown to the agent. All relevant code, data, tools, and assets are available through the terminal, and the agent may inspect files, edit programs, run commands, and evaluate intermediate outputs until it succeeds or reaches its time budget.

The benchmark is long-horizon by construction. A task requires continued work over many dependent steps rather than a single patch or answer: reproducing an experiment, repairing a scientific analysis pipeline, optimizing a system, solving a campaign of game levels, or inferring the behavior of an unknown artifact. Errors made early can propagate through later stages, and apparent completion must be checked against an end-to-end objective.

\subsection{Benchmark extension and splits}

\oursabbr{} contains all 46 tasks from \lhtbone and 31 additions. We preserve the original task identities and descriptions so that the extension remains traceable to the first benchmark, while repairing environment and verifier integration defects discovered during larger evaluations. The 77 tasks are organized into two complementary splits:
\begin{itemize}
    \item \textbf{Released-workflow split (46 tasks).} The 46 original tasks and one additional optimization task cover audit and repair, optimization, gameplay, experiment reproduction, constraint puzzles, professional workflows, and engineering. Their graders directly expose dense subtask or continuous rewards.
    \item \textbf{Reconstruction split (31 tasks).} These tasks ask an agent to reproduce the behavior of a parameterized scientific or engineering artifact from public observations. 
Hidden scenarios test generalization, and a common six-stage ladder converts verifier evidence into dense progress rewards.
\end{itemize}

This split separates two sources of long-horizon difficulty. Released-workflow tasks require navigating a large, heterogeneous project whose intended objective is stated but difficult to complete. Reconstruction tasks make the implementation target itself partially latent: agents must infer behavior from evidence and then implement a solution that generalizes to withheld cases. 

\paragraph{Dense reward and completion.}
We use the same normalized scale for both benchmark splits. For reconstruction tasks, the dense ladder score replaces the sparse final binary indicator in all cross-task results. We report mean $R$, a soft pass at $R\geq0.95$, and a strict pass at $R=1.0$. Mean reward captures partial progress; the two thresholds distinguish nearly complete solutions from exact completion.




\subsection{Evaluation hardening}
%
We extend each task run to three hours and invoke the verifier every 30 minutes to track progress. 
These checkpoints can reveal agent regressions, such as overwriting previously correct work, as well as evaluation-system defects, including stale verifier artifacts and hidden grader data leaking across phases---problems that a single end-of-run score would not expose.
\oursabbr{} therefore hardens both task execution and score collection. 
We fixed three harness/configuration defects that prevented tasks using separate verifier environments from producing rewards, isolated verifier phases from the active agent process tree, cleared verifier logs before resuming the agent, and audited successful trajectories for access to hidden fixtures or grader artifacts. 

\subsection{Domains}

\begin{figure}[H]
  \centering
  \includegraphics[width=0.90\linewidth]{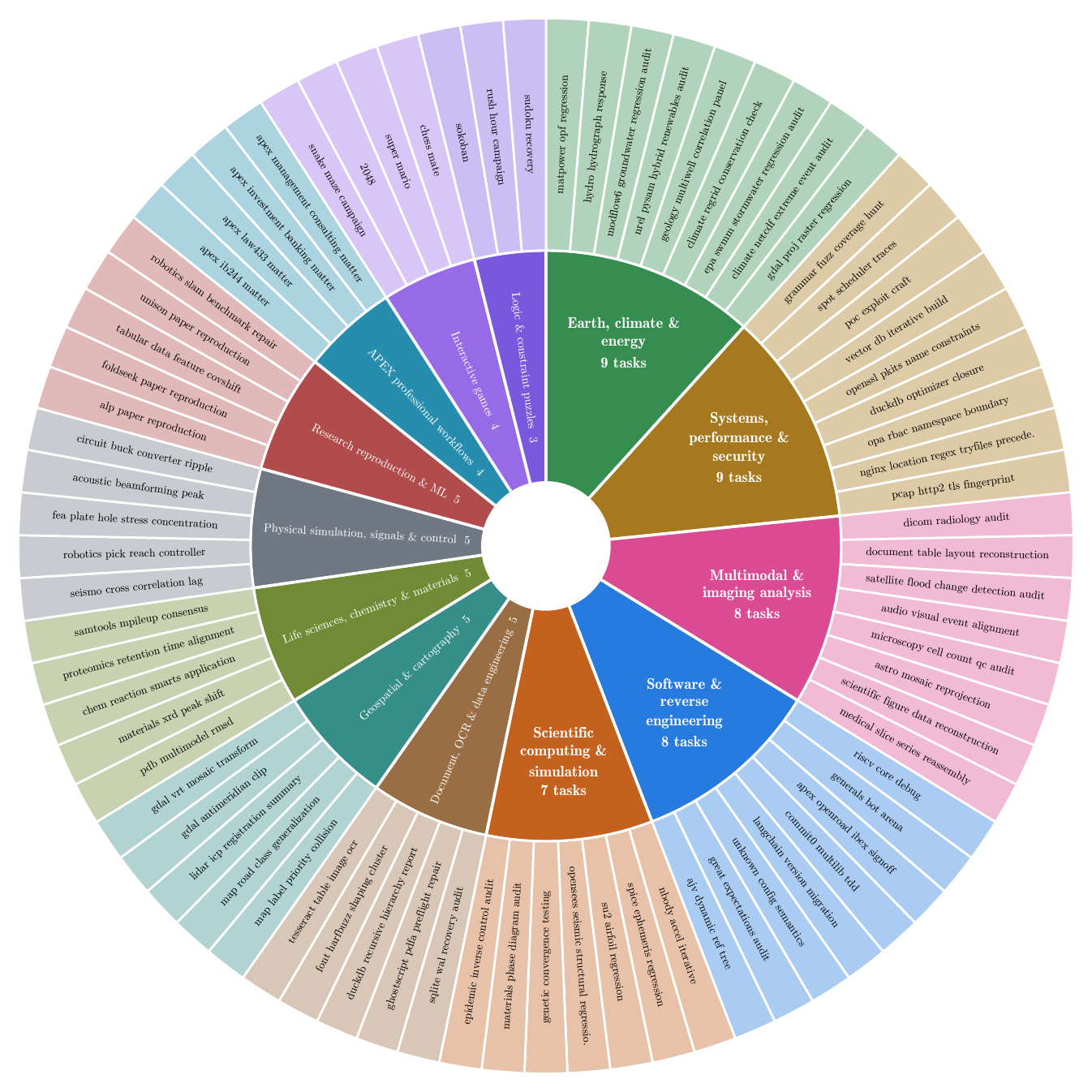}
  \caption{\textbf{Task distribution across the 13 \oursabbr{} domains.}
  The 77 tasks combine the original benchmark domains with geospatial and cartographic workflows; document, OCR, and data engineering; life sciences, chemistry, and materials; and physical simulation, signals, and control.}
  \label{fig:lhtb-categories}
\end{figure}

The benchmark spans 13 domains: APEX professional workflows~\cite{vidgen2026apexagents}; earth, climate, and energy; interactive games; logic and constraint puzzles; multimodal and imaging analysis; research reproduction and machine learning; scientific computing and simulation; software and reverse engineering; systems, performance, and security; geospatial and cartographic workflows; document, OCR, and data engineering; life sciences, chemistry, and materials; and physical simulation, signals, and control. Figure~\ref{fig:lhtb-categories} summarizes their distribution, while Appendix~\ref{app:task-list} provides the full 77-task inventory.

\section{Task Construction Examples}
\label{sec:construction}
The original LHTB tasks were individually designed around complete workflows. 
This produces realistic and varied environments, but it is costly to scale: every new task requires a plausible project, a reproducible solution, robust hidden tests, and a dense grader. 
\oursabbr{} retains that design for the released-workflow split and introduces a complementary construction procedure that turns working domain artifacts into long-horizon reconstruction tasks.

\begin{figure}[p]
  \centering
  \textbf{(a) LiDAR registration and point-cloud summaries}\par
  \includegraphics[width=0.89\linewidth]{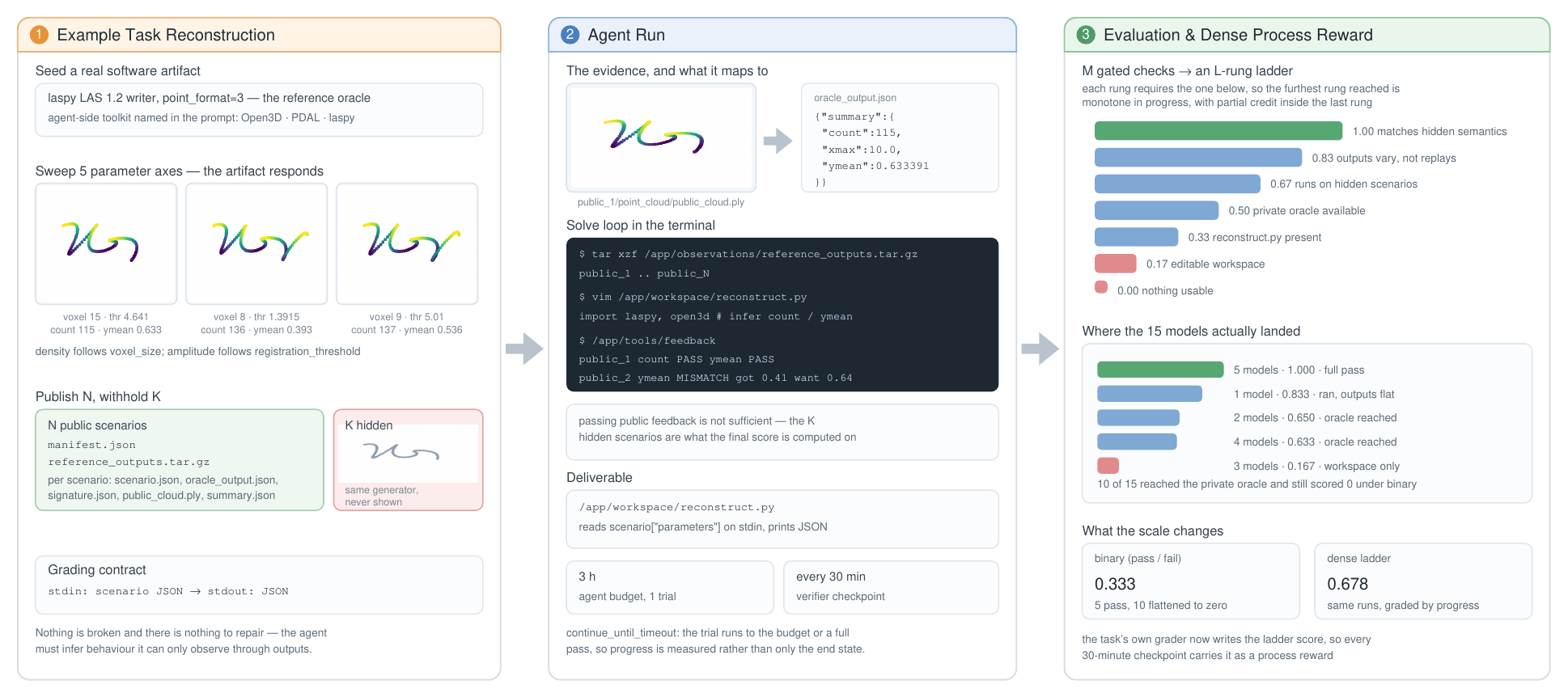}
  \vspace{0.4em}

  \textbf{(b) Chemical reaction templates and molecular descriptors}\par
  \includegraphics[width=0.89\linewidth]{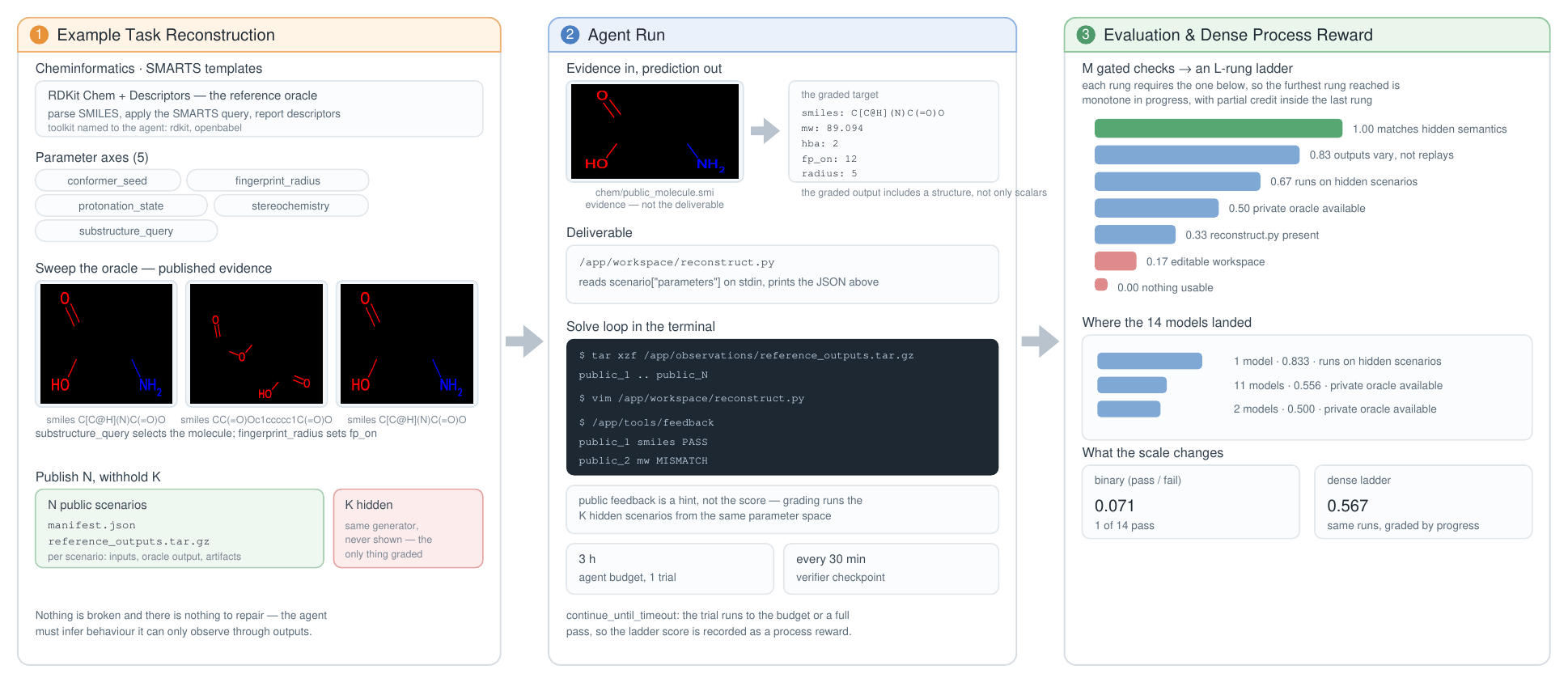}
  \vspace{0.4em}

  \textbf{(c) Acoustic beamforming and room simulation}\par
  \includegraphics[width=0.89\linewidth]{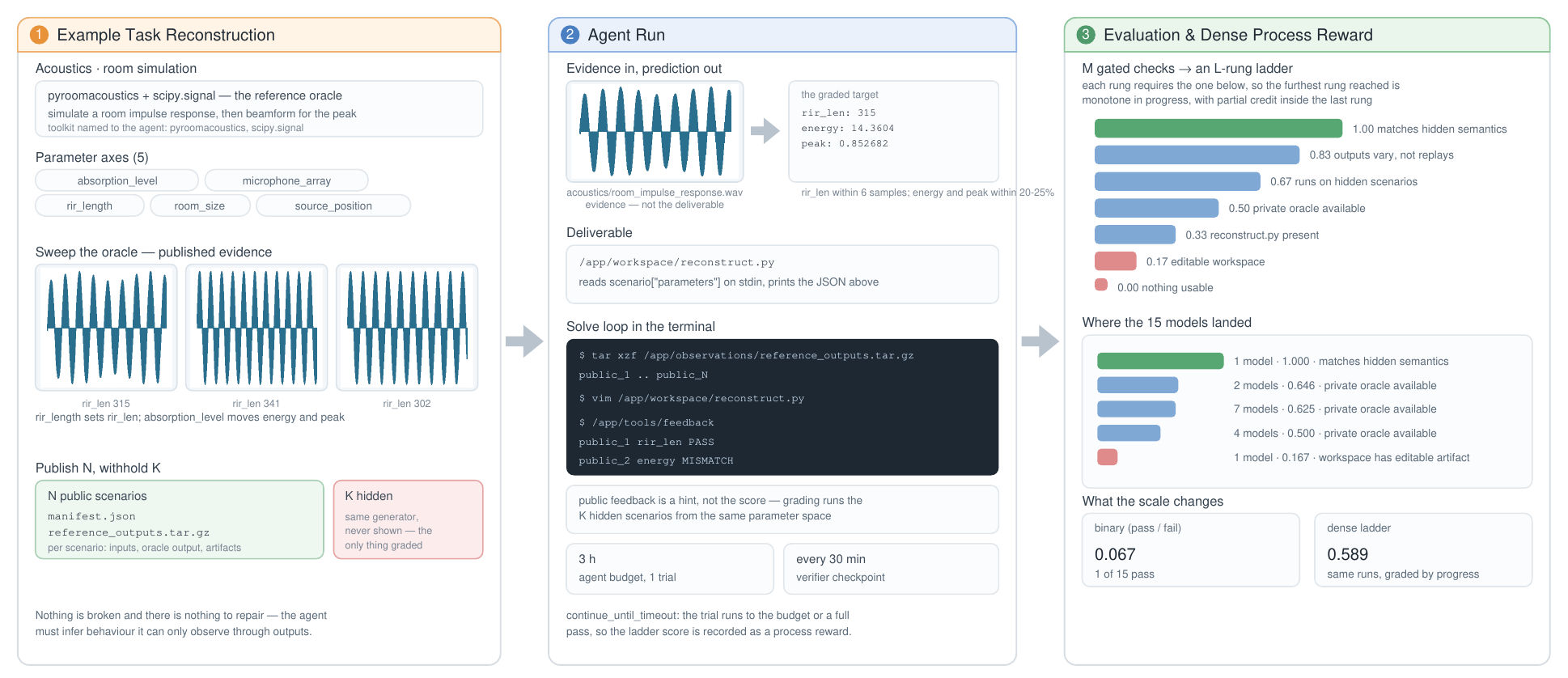}
  \caption{\textbf{Construction of reconstruction tasks.} A working reference artifact is swept over meaningful parameter axes. $N$ scenarios are exposed as behavioral evidence and $K$ are withheld for grading. The agent must implement a terminal program that reproduces behavior on unseen scenarios; public feedback guides debugging but does not reveal the hidden score. The same contract applies to point clouds (a), molecules (b), and waveforms (c).}
  \label{fig:construction-examples}
\end{figure}

\subsection{Construction examples}

Figure~\ref{fig:construction-examples} shows three instances of our task construction recipe. 
The visual artifacts are the actual public evidence supplied to agents. 
LiDAR point clouds expose changes in density and amplitude, molecular diagrams expose reaction and descriptor behavior, and acoustic waveforms expose changes in simulated impulse responses. 
Each task uses the same public/hidden split, terminal interface, anti-memorization gate, and checkpointed evaluation protocol.

\subsection{Task types}

The construction pipeline broadens the benchmark, where \oursabbr{} contains eight task types:

\begin{table}[H]
\centering
\small
\setlength{\tabcolsep}{5pt}
\begin{tabular}{p{0.24\linewidth}r p{0.60\linewidth}}
\toprule
\textbf{Task type} & \textbf{Count} & \textbf{Long-horizon objective} \\
\midrule
Reconstruction & 30 & Infer and reproduce a reference artifact's behavior from public observations. \\
Audit and repair & 23 & Diagnose and repair a broken analysis or audit pipeline against hidden cases. \\
Optimization & 6 & Iteratively improve a working implementation toward quality or performance targets. \\
Gameplay & 5 & Act over a long interactive campaign. \\
Reproduction & 4 & Reproduce a published experiment, benchmark, or scientific result. \\
Professional workflow & 4 & Complete a multi-stage professional matter with deterministic grading. \\
Constraint puzzle & 3 & Recover a legal solution through iterative interaction with a puzzle environment. \\
Engineering signoff & 2 & Drive a hardware or systems flow until all acceptance gates pass. \\
\bottomrule
\end{tabular}
\caption{\textbf{Task types in \oursabbr{}.} Reconstruction is the new automated construction family; the remaining types preserve the heterogeneous workflow structure of the original benchmark.}
\label{tab:task-types}
\end{table}

\subsection{Quality control}

Every task must satisfy three checks before inclusion. First, the reference solution must run reproducibly and achieve $R = 1.0$ on all hidden cases. 
Second, the starter materials should be incomplete enough that solving the task requires genuine work, rather than a trivial patch or simply copying from what is already visible. 
Third, we use DeepSeek-V4-Pro~\cite{deepseekai2026deepseekv4} to review alignment between each task's instructions and verifier, ensuring that the instructions provide enough information to satisfy the grader without revealing hidden answers. 
%

\section{Experiments}
\label{sec:experiments}

\paragraph{Harness and agents.}
We evaluate models using the Harbor framework with the Terminus-2 agent harness~\cite{Harbor_Framework}, which interacts with each task environment through a single long-horizon terminal session.
%
%

\paragraph{Models.}
We benchmark GPT-6-astra~\cite{openai2026gpt6astra}, GPT-5.6-sol~\cite{openai2026gpt56sol}, DeepSeek V4 Pro~\cite{deepseekai2026deepseekv4}, DeepSeek V4 Flash~\cite{deepseekai2026deepseekv4}, Gemini 3.1 Pro~\cite{google2026gemini31pro}, Gemini 3.8 Flash~\cite{google2026gemini38flash}, GLM-5.3~\cite{zai2026glm53}, GLM-5.2~\cite{zai2026glm52}, Kimi K3~\cite{kimiteam2026kimik3}, Kimi K2.6~\cite{moonshot2026kimik26}, MiniMax M3~\cite{minimax2026minimaxm3}, Qwen 3.8 Max~\cite{alibaba2026qwen38max}, Doubao Seed 2.1 Pro~\cite{bytedance2026seed21}, Hunyuan hy4~\cite{tencent2026hy4preview}, and Grok-4.6~\cite{xai2026grok46}.\footnote{All models run with xhigh reasoning.}


\paragraph{Metrics.}
For each model, we report the soft pass rate at $R \geq 0.95$, the strict pass rate (the fraction of tasks with normalized reward exactly $1.0$), the mean normalized reward across all tasks, the number of episodes required per task, the elapsed runtime per task, the estimated dollar cost per task.

\subsection{Main Results}

%

\begin{figure}[t]
  \centering
  \includegraphics[width=\linewidth]{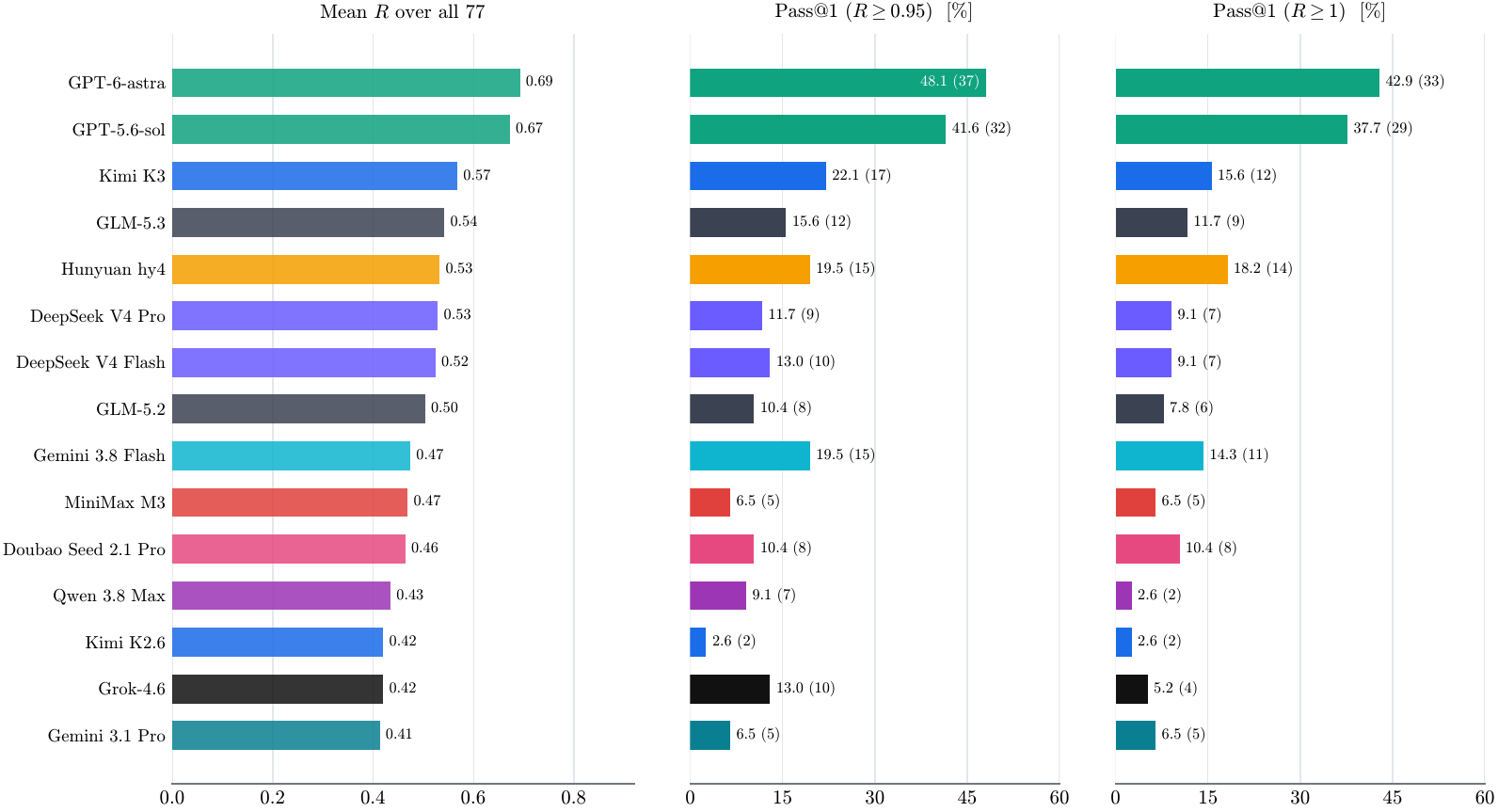}
  \caption{\textbf{There exists a gap between solve rate and process reward, where models can achieve significant progresses on overall tasks but fail most of them.} Pass rate is the fraction of the 77 tasks whose overall reward $R$ meets a threshold $\tau$, with $R \geq 1.0$ corresponding to full completion, while mean $R$ captures partial progress across all tasks. GPT-6-astra achieves the strongest overall performance, resolving 48.1\% of tasks at $R \geq 0.95$ and 42.9\% at $R = 1.0$.}
  \label{fig:lhtb-leaderboard}
\end{figure}

We set the timeout limit for each task to be 3 hours for cost reasons.
Figure~\ref{fig:lhtb-leaderboard} summarizes pass rate at $R \geq 0.95$ and $R \geq 1.0$, along with mean reward across all 77 tasks.
Overall, the evaluated frontier models still struggle on long-horizon tasks, even under a dense-grading evaluation that gives credit for partial progress.
GPT-6-astra is the strongest reported model, but still resolves only 48.1\% of tasks (37/77) at $R \geq 0.95$. GPT-5.6-sol follows at 41.6\% (32/77). Kimi K3 is a distant third at 22.1\% (17/77). A second tier includes hy4 and Gemini 3.8 Flash at 19.5\% (15/77) each, followed by GLM-5.3 at 15.6\% (12/77).
The gap between pass rate and mean reward (0.2-0.5) for several mid-tier systems indicates that many models make substantial partial progress on hard tasks without fully resolving them.
The results suggest that long-horizon execution is not merely a local reasoning problem, but remains a central bottleneck even for top frontier models.

\paragraph{GPT-6-astra takes the lead on software reconstruction tasks.}
GPT-6-astra's overall lead is concentrated in the reconstruction split. On the 30 reconstruction tasks it reaches a mean reward of $0.93$ and solves 24 tasks at $R \geq 0.95$, ahead of GPT-5.6-sol ($0.84$, 18 tasks) and far ahead of every other model ($\leq 0.71$, at most 9 tasks).
On the 47 released-workflow tasks, GPT-5.6-sol is slightly ahead ($0.57$, 14 tasks), with GPT-6-astra second ($0.54$, 13 tasks) and DeepSeek V4 Flash close behind ($0.53$).
GPT-6-astra is also more step-efficient on reconstruction tasks, solving them after a median of 37 agent steps and 26 minutes, against 67 steps and 36 minutes for GPT-5.6-sol.
In the first 40 steps of runs that last that long, about one third of GPT-6-astra's commands write or edit files and only $28\%$ repeat an earlier command; for GPT-5.6-sol the corresponding numbers are $14\%$ and $50\%$.
On released tasks, that speed buys less: GPT-6-astra is close to its final reward after one hour ($0.53$ at 60 minutes, $0.56$ at three hours), while GPT-5.6-sol keeps improving ($0.53$ to $0.58$) and finishes ahead.
Over all 77 tasks GPT-6-astra remains ahead throughout, although GPT-5.6-sol closes most of the gap by the end of the run (Section~\ref{sec:analysis-horizon}).

\paragraph{Open-source models, model performance broadly tracks release date}.
Earlier releases such as Kimi K2.6 (April 20, 2026) and GLM-5.2 (June 16, 2026) are surpassed by later releases such as Kimi K3 (July 16, 2026; open weights July 27), GLM-5.3 (August 14, 2026; open weights August 28), and Hunyuan hy4 preview (August 28, 2026).

\subsection{Success and Failures}
\label{sec:analysis-domains}

\begin{figure}[t]
  \centering
  \includegraphics[width=\linewidth]{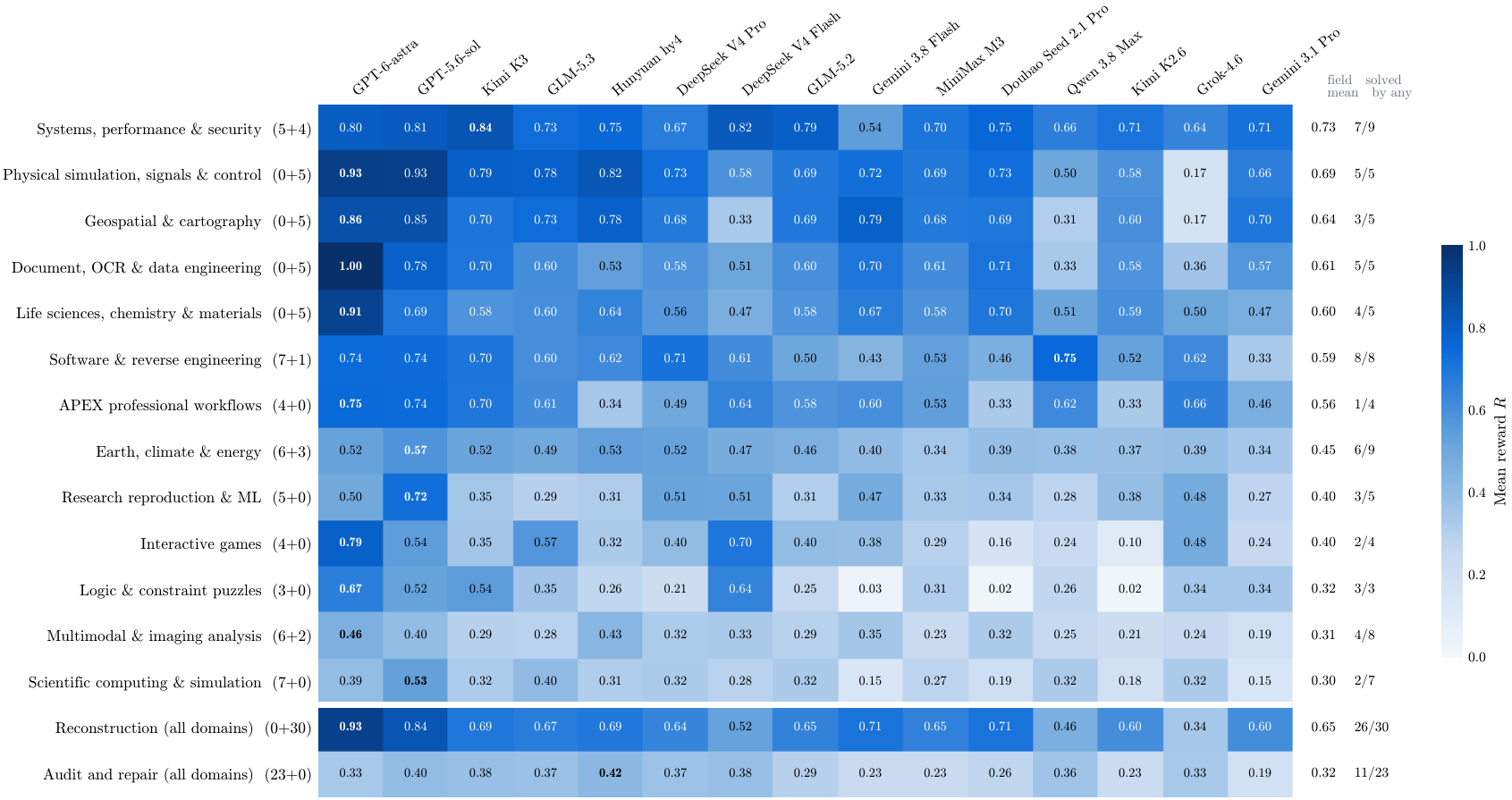}
  \caption{\textbf{Mean reward by category and model.}
  Rows are the 13 categories, sorted by the mean over all models, followed by the two most common task types pooled across domains; parentheses give the number of released (from LHTB-1.0) and reconstruction tasks (newly added). Each cell averages the dense reward over the row's tasks, counting unscored tasks as $0$, and the best model in each row is shown in bold. The right-hand columns give the mean over all models and the number of tasks that at least one model solves ($R \geq 0.95$).}
  \label{fig:lhtb-domains}
\end{figure}

Figure~\ref{fig:lhtb-domains} breaks mean reward down by category, task type, and model.
\paragraph{Software and systems tasks are the most tractable for every model.}
Systems, performance \& security has the highest mean across models ($0.73$), every software \& reverse-engineering task is solved by at least one model, and the most widely solved task, \texttt{riscv-core-debug}, is solved by 13 of the 15 models.
GPT-6-astra is strong but not dominant: it ranks second on software \& reverse engineering, just behind Qwen 3.8 Max ($0.74$ vs.\ $0.75$), and fourth on systems, where Kimi K3 leads ($0.84$).
Its clearest advantages are on interactive games ($0.79$ against a fleet mean of $0.40$) and on document, OCR \& data engineering, where it solves all five tasks.

\paragraph{Scientific tasks are among the hardest overall.}
Scientific computing \& simulation ($0.30$) and multimodal \& imaging analysis ($0.31$) are the two lowest-scoring categories.
Of the 24 tasks that no model solves, 20 come from the released split and 12 are audit-and-repair tasks; on nine tasks, no model exceeds $R = 0.2$, and eight of those nine are audit-and-repair.
The main difficulty is therefore sustained debugging and repair, not a single missing final step.

%
%

\subsection{Solve Rate and Reward over Time and Steps}
\label{sec:analysis-horizon}

\begin{figure}[t]
  \centering
  \includegraphics[width=\linewidth]{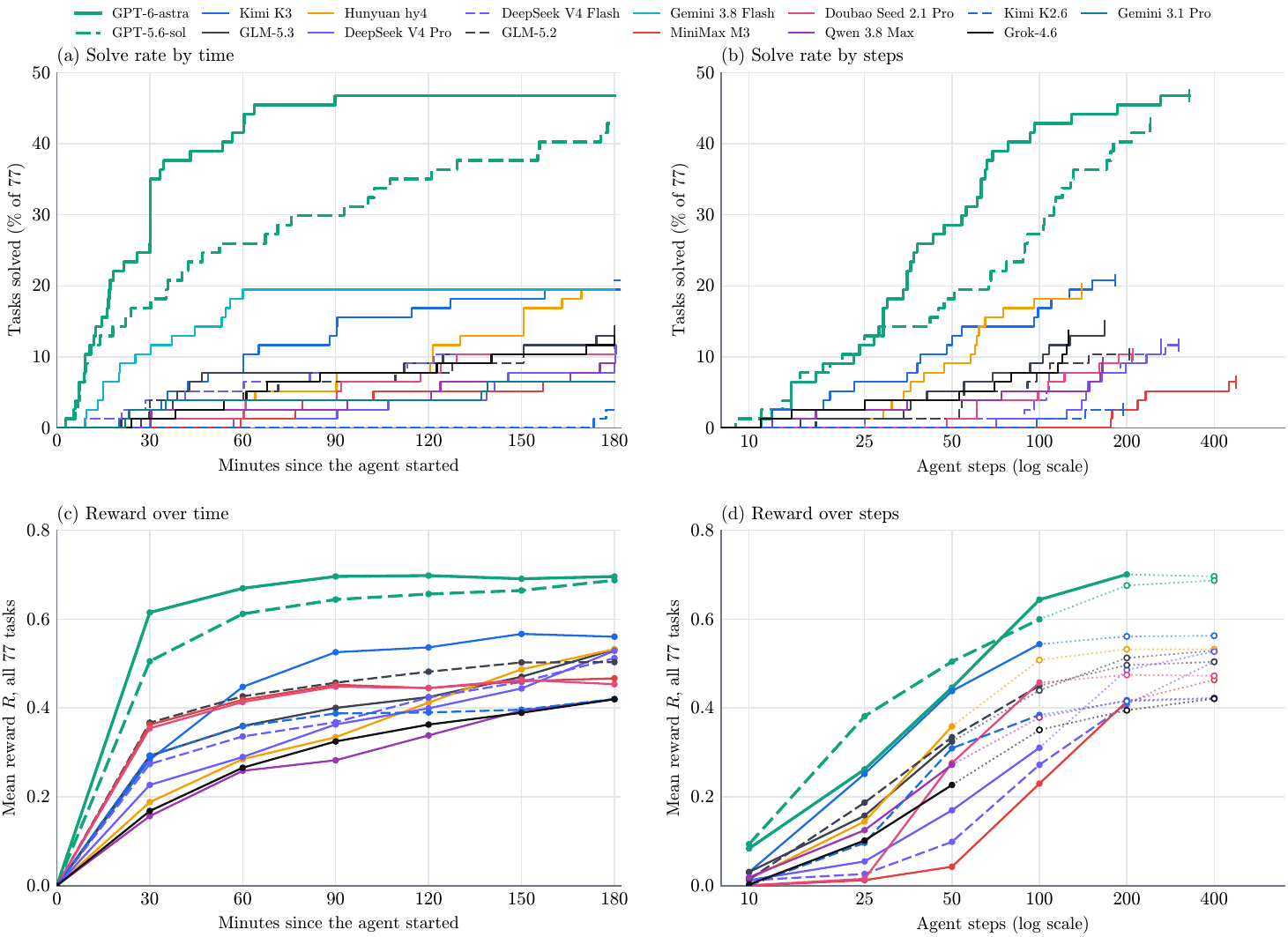}
  \caption{\textbf{Solve rate and reward over the three-hour run.}
  (a)~Many solves arrive after the first hour. Share of the 77 tasks solved by each minute; a task counts as solved at the first checkpoint whose verifier reward reaches $0.95$.
  (b)~Late wall-clock solves can reflect slow steps. The same solve curves against agent steps; each curve ends at the median number of steps the model completes in three hours.
  (c)~Mean reward keeps rising after the first hour.
  (d)~More steps do not guarantee more reward. The same mean reward against agent steps. Hollow markers and dotted segments mark budgets that more than $20\%$ of a model's runs never reach.}
  \label{fig:lhtb-horizon}
\end{figure}

We follow each model through the three-hour budget in both wall-clock time and agent steps (Figure~\ref{fig:lhtb-horizon}).
A task counts as softly solved at the first checkpoint whose verifier reward reaches $0.95$, which gives 189 solves over 1,143 runs and differs from the dense leaderboard by at most one task per model.

\paragraph{The three-hour budget changes the middle of the leaderboard.}
Across models, progress often continues well beyond the first hour: the median successful run solves its task after 57 minutes, with $42\%$ of all solves appearing after the one-hour checkpoint and $21\%$ after the two-hour checkpoint.
The top stabilizes early. GPT-6-astra solves 27 tasks within 30 minutes and 34 of its 36 solved tasks within the first hour, with a median solve time of 24 minutes.
The middle does not. Hunyuan hy4 (median 121 minutes) and DeepSeek V4 Pro (141 minutes) solve many tasks late, so a one-hour evaluation would rank hy4 ninth instead of fourth and DeepSeek V4 Pro 13th instead of eighth.

\begin{figure}[tb]
  \centering
  \includegraphics[width=\linewidth]{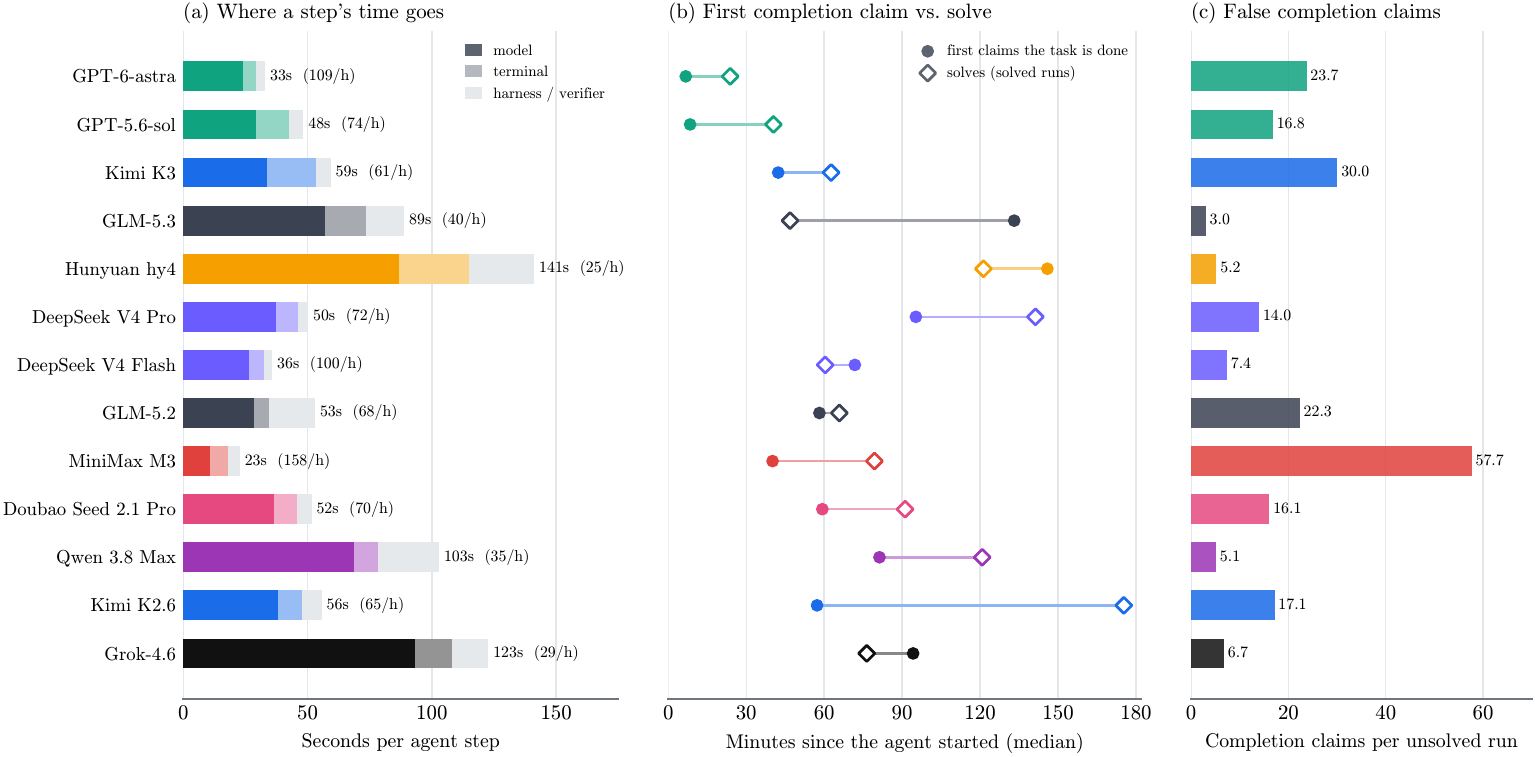}
  \caption{\textbf{Wall-clock performance reflects both step speed and overconfident stopping.}
  (a)~Some models spend much of each step waiting for the model, so they take far fewer steps within three hours.
  (b)~Fast models tend to claim completion earlier, while slower models often wait much longer before doing so.
  (c)~Many early completion claims are false alarms, especially for the most aggressive models.}
  \label{fig:lhtb-runtime}
\end{figure}

\paragraph{Step efficiency and step latency tell different stories.}
Wall-clock time combines how many agent steps a task requires and how long each step takes.
Step latency alone varies by roughly $6\times$, from 23 seconds per step for MiniMax M3 to 141 seconds for Hunyuan hy4; across models, agents spend $48$--$76\%$ of each run waiting for the model rather than interacting with the environment (Figure~\ref{fig:lhtb-runtime}).
This changes how late solves should be interpreted. Hy4, for example, needs a median of only 58 steps to solve a task, fewer than GPT-5.6-sol's 68, yet it completes fewer than 80 steps within the three-hour budget.
By contrast, the most step-hungry models are DeepSeek V4 Pro and MiniMax M3, which require medians of 147 and 218 steps to solve a task.
Under an equal-step comparison, GPT-6-astra remains clearly best: it solves 22 tasks within 50 steps, compared with 14 for GPT-5.6-sol and at most 10 for any other model.



\paragraph{Fast models also overclaim completion.}
Completion claims are informative because the harness verifies them and resumes the run on failure.
GPT-6-astra and GPT-5.6-sol make their first completion claim after only 7 and 8 minutes on median, and on unsolved runs still claim completion 24 and 17 times on average; MiniMax M3 does so 58 times.
By contrast, Hunyuan hy4, GLM-5.3, Grok-4.6, and DeepSeek V4 Pro often delay their first claim until 1.5--2.5 hours and average only 3--14 claims per unsolved run (Figure~\ref{fig:lhtb-runtime}).
Across the fleet, $92\%$ of completion claims occur on runs that never pass verification.

\subsection{Time and Process Reward Analysis}
\label{sec:analysis-false-finish}

\begin{figure}[t]
  \centering
  \begin{minipage}[t]{0.49\linewidth}
    \centering
    \includegraphics[width=\linewidth]{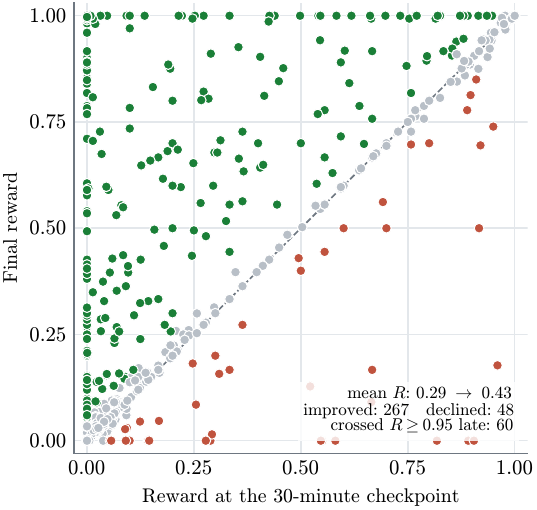}

    \vspace{0.2em}
    {\small \textbf{30-minute checkpoint}}
  \end{minipage}\hfill
  \begin{minipage}[t]{0.49\linewidth}
    \centering
    \includegraphics[width=\linewidth]{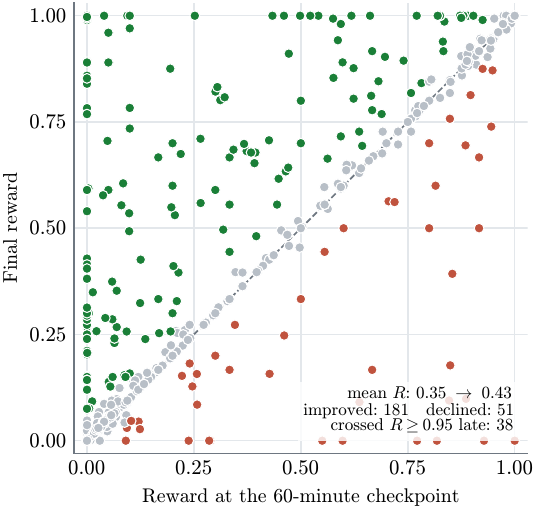}

    \vspace{0.2em}
    {\small \textbf{60-minute checkpoint}}
  \end{minipage}
  \caption{\textbf{Many runs continue making meaningful progress well after 30 and 60 minutes.}
  Each point is one scored task run: the x-axis shows reward at the 30-minute checkpoint (left) or 60-minute checkpoint (right), and the y-axis shows final reward. The dashed diagonal marks no change. Green points improve by more than $0.05$, red points decline by more than $0.05$, and gray points stay within that band. In many runs, reward continues to increase after the checkpoint, and some runs newly cross the soft-pass threshold only later.}
  \label{fig:lhtb-time-ablation}
\end{figure}

\begin{figure}[t]
  \centering
  \includegraphics[width=0.78\linewidth]{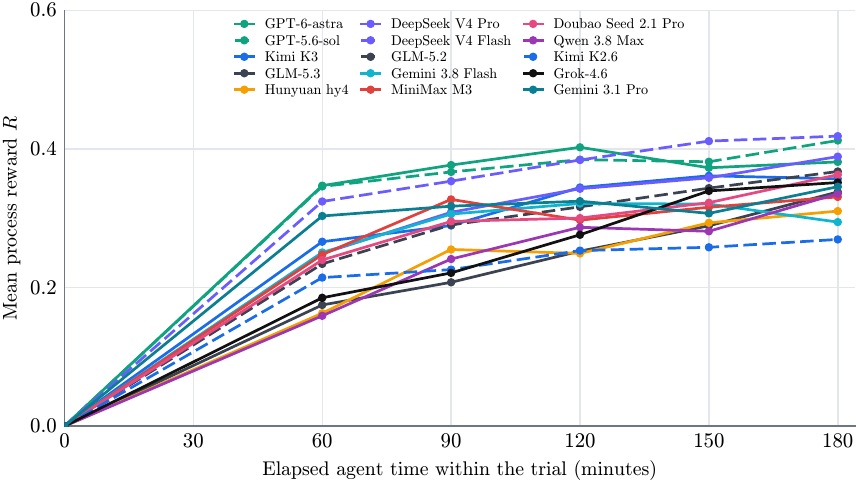}
  \caption{\textbf{Even unsolved runs continue making progress for hours.}
  Each curve includes only released-task runs whose final reward remains below $0.95$. Curves start at $(0,0)$ by definition and omit the sparsely populated 30-minute interval. Within each time bin, checkpoints are first averaged per task and then across tasks, so every task receives equal weight regardless of how many checkpoints it produced.}
  \label{fig:lhtb-process-reward}
\end{figure}

\paragraph{Process rewards let us measure partial progress during a long-horizon run, not just the final outcome.}
Figure~\ref{fig:lhtb-time-ablation} shows that many runs continue improving after the first checkpoint. Mean reward rises from $0.29$ at 30 minutes to $0.43$ by the end of the run; 267 of 693 runs gain more than $0.05$ after 30 minutes, and 60 cross the soft-pass threshold only after that checkpoint. Even after 60 minutes, 181 runs still gain more than $0.05$, and 38 later cross the threshold.

\paragraph{Regression could exist but is not the dominant pattern.} 
We find 65 model--task trajectories whose final reward is more than $0.05$ below either the 30 or 60-minute checkpoint; 55 continue declining later, and 54 end at the time limit. 
These cases are concentrated in games or puzzles, where later actions can damage state or overwrite useful outputs.

\paragraph{Most unsolved shows that unsolved runs still make progress.} After removing successful runs, every model's mean reward at 180 minutes remains above its 60-minute level, and the stronger models maintain higher process rewards throughout (Figure~\ref{fig:lhtb-process-reward}). 
Many failures therefore reflect incomplete long-horizon progress rather than immediate stagnation.


\section{Cost Analysis}
\label{app:cost-analysis}

\begin{table}[H]
  \centering
  \small
  \setlength{\tabcolsep}{4pt}
  \begin{tabular}{@{}lrrrr@{}}
    \toprule
    \textbf{Model} & \textbf{Tokens / task (M)} & \textbf{LLM calls / task} & \textbf{LLM time / task (min)} & \textbf{Cost / task (\$)} \\
    \midrule
    Hunyuan hy4                     &    4.4 &   78.9 &  102.7 &    3.71 \\
    Doubao Seed 2.1 Pro &   18.0 &  200.2 &  111.7 &    9.12 \\
    DeepSeek V4 Flash               &   45.5 &  290.1 &  127.0 &   10.07 \\
    GLM-5.2                         &   27.9 &  214.0 &   93.0 &   15.43 \\
    Qwen 3.8 Max                    &    8.8 &  113.3 &  119.0 &   17.67 \\
    DeepSeek V4 Pro                 &   31.2 &  231.0 &  130.8 &   18.21 \\
    Kimi K2.6                       &   19.5 &  191.6 &  123.2 &   18.70 \\
    GLM-5.3                         &   20.5 &  137.3 &  110.8 &   18.83 \\
    MiniMax M3                      &   71.0 &  452.5 &   84.3 &   21.44 \\
    Grok-4.6                        &   12.8 &  133.3 &  132.8 &   25.91 \\
    Kimi K3                         &   16.2 &  170.8 &   93.7 &   32.09 \\
    Gemini 3.1 Pro                  &   13.3 &  500.2 &     9.5 &   32.93 \\
    Gemini 3.8 Flash                &   48.2 &  836.4 &     12.7 &   37.31 \\
    GPT-5.6-sol                     &   33.5 &  175.0 &   80.1 &   67.60 \\
    GPT-6-astra                     &   39.4 &  213.5 &   83.2 &  397.16 \\
    \midrule
    \textit{Average}                &   27.3 &  262.5 &   94.3 &   48.41 \\
    \bottomrule
  \end{tabular}
  \setlength{\tabcolsep}{6pt}
  \caption{\textbf{Cost varies sharply across models, and the strongest model is by far the most expensive.}
  On average, models spend 262.5 LLM calls and 94.3 minutes per task, at an estimated \$48.4 per task.
  Costs are estimated from per-task token usage and public list prices.
  Tokens/task counts both input and output tokens, and Figure~\ref{fig:lhtb-cost-reward} uses the corresponding 77-task suite totals.
  $^{\dagger}$ priced from the nearest listing rather than an exact match.}
  \label{tab:lhtb-cost}
\end{table}

\begin{figure}[tb]
  \centering
  \begin{minipage}[t]{0.49\linewidth}
    \centering
    \includegraphics[width=\linewidth]{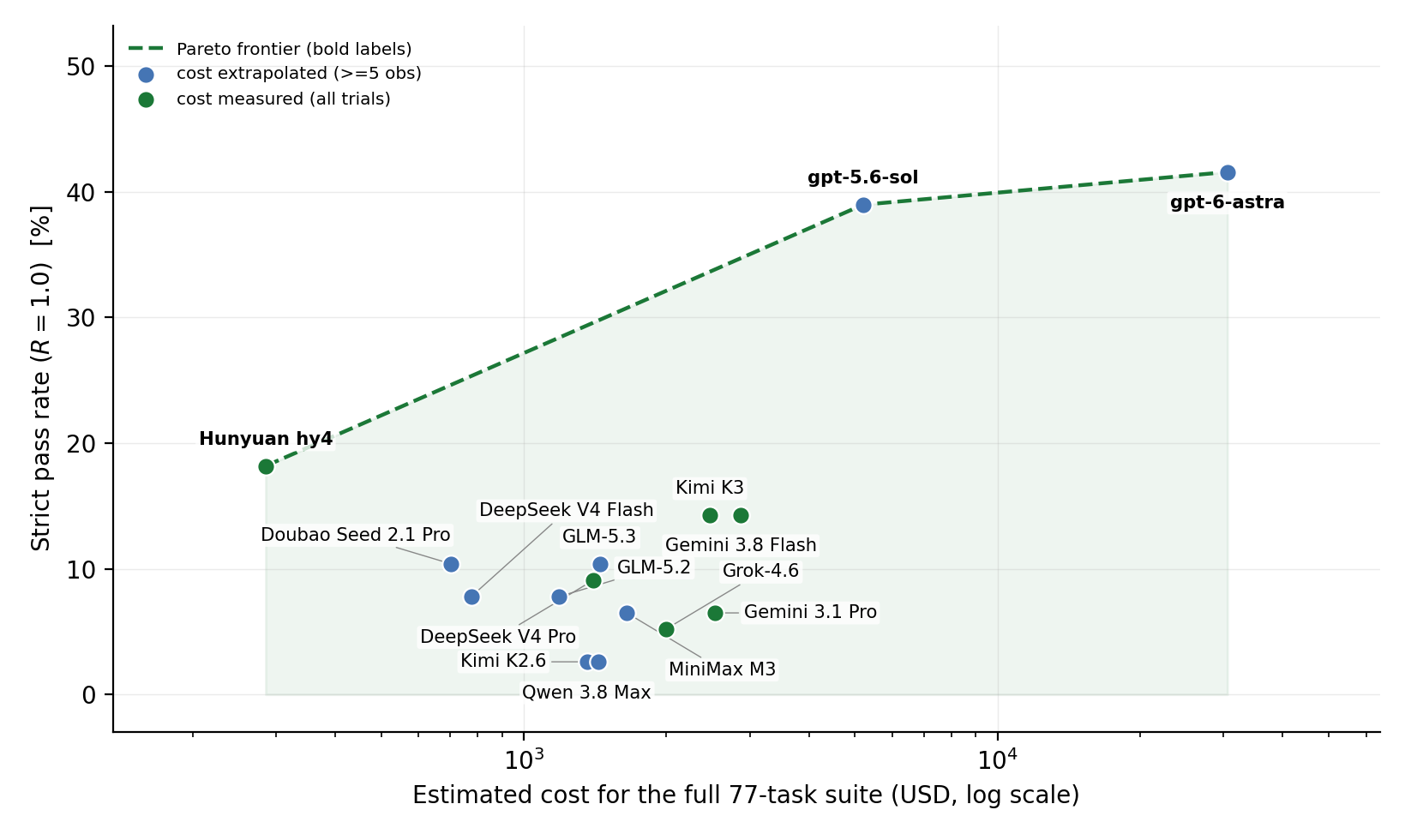}

    \vspace{0.2em}
    {\small \textbf{Strict pass rate ($R = 1.0$)}}
  \end{minipage}\hfill
  \begin{minipage}[t]{0.49\linewidth}
    \centering
    \includegraphics[width=\linewidth]{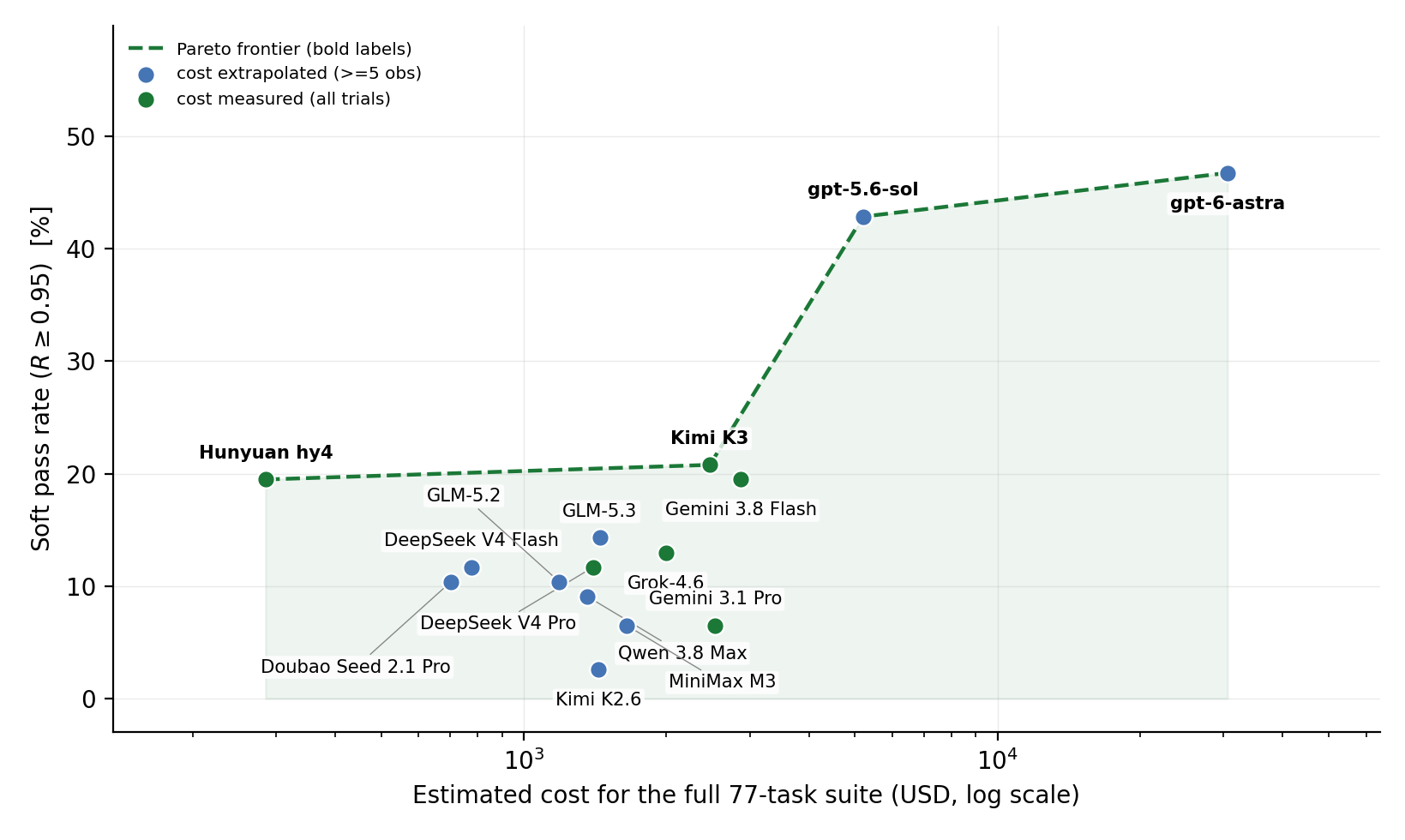}

    \vspace{0.2em}
    {\small \textbf{Soft pass rate ($R \geq 0.95$)}}
  \end{minipage}
  \caption{\textbf{Better long-horizon performance comes at sharply higher cost.}
  Estimated cost of the full 77-task suite (log scale) against strict pass rate ($R = 1.0$, left) and soft pass rate ($R \geq 0.95$, right). The dashed line traces the Pareto frontier, and bold labels mark models on that frontier. Softening the pass criterion adds Kimi K3 to the frontier. Hunyuan hy4 is the strongest low-cost model, while GPT-6-astra is the strongest overall at the highest cost.}
  \label{fig:lhtb-cost-reward}
\end{figure}

We report the cost estimates reconstructed from per-task token usage and public list prices (Table~\ref{tab:lhtb-cost}), which Figure~\ref{fig:lhtb-cost-reward} summarizes with side-by-side Pareto cost--reward frontiers under strict and soft pass criteria.
The estimated cost of running the full benchmark spans from about \$285 for Hunyuan hy4 to about \$30{,}581 for GPT-6-astra.
Although GPT-6-astra achieves the highest pass rate, it does so at by far the highest total cost. 
GPT-5.6-sol forms a second tier of strong but substantially cheaper performance at roughly one sixth of that cost, while Kimi K3 sits further down the cost curve with much lower spend but also a noticeably lower resolution rate.
Hunyuan hy4 anchors the low-cost end of the Pareto frontier, while GPT-5.6-sol and GPT-6-astra extend that frontier upward as progressively stronger but more expensive models. 
Under the soft $R \geq 0.95$ criterion, Kimi K3 joins the frontier because several of its near-misses are counted as passes rather than perfect completions. 
%

\begin{figure}[!tb]
  \centering
  \includegraphics[width=\linewidth]{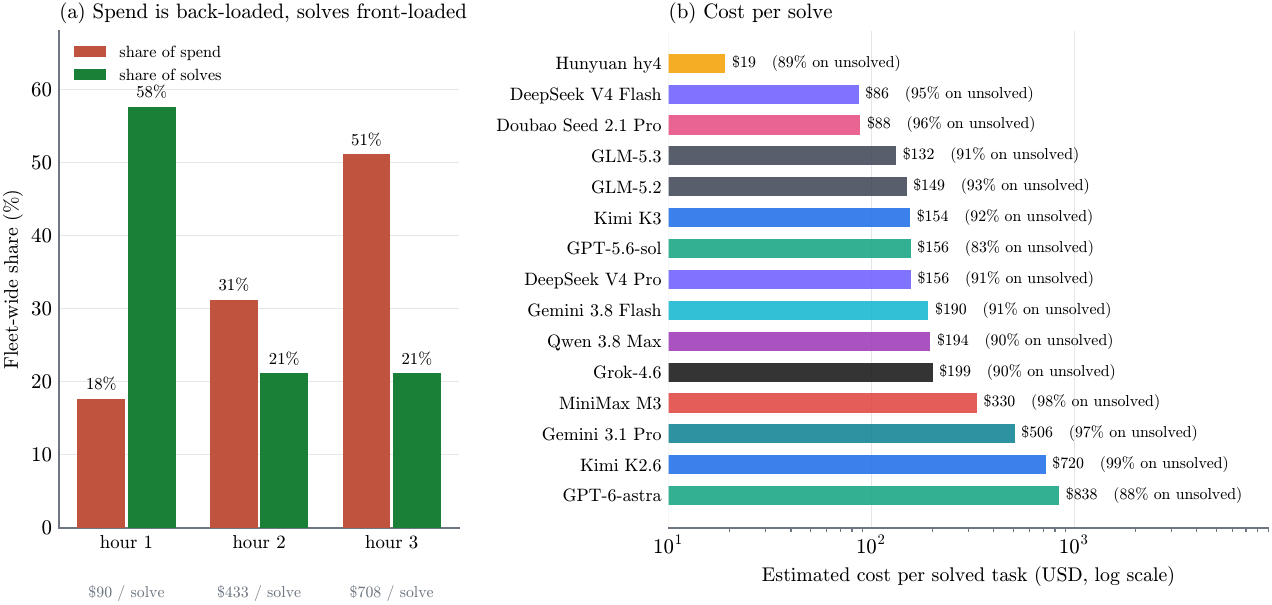}
  \caption{\textbf{Most spending goes to the last hour and to tasks that never solve.}
  (a)~Solves are front-loaded, but spending is back-loaded. This panel shows the fleet-wide share of estimated spend and of solves in each hour of the run; below each pair, the spend in that hour divided by the solves found in it.
  (b)~Most model spending goes to tasks that never solve. This panel shows estimated cost per solved task for each model (log scale), with the share of the model's spend that goes to tasks it never solves.
  Costs use public list prices without prompt-cache discounts.}
  \label{fig:lhtb-cost-hour}
\end{figure}

\paragraph{Most spending goes to unsolved tasks and to the last hour.}
Figure~\ref{fig:lhtb-cost-hour} breaks spending down by outcome and by hour.
Between $83\%$ and $99\%$ of every model's spend goes to tasks it never solves; GPT-6-astra spends an average of \$664 on an unsolved task and \$101 on a solved one.
This is due to that every step resends a longer history, so hours one, two, and three account for $18\%$, $31\%$, and $51\%$ of benchmark-wide spend, while solves are front-loaded ($58\%$, $21\%$, and $21\%$).
A solve found in the third hour therefore costs about eight times as much as one found in the first (\$708 against \$90).\footnote{Our estimates apply no prompt-cache discount, which would cheapen the resent history, so this ratio is an upper bound.}
Per solved task, costs range from \$19 for Hunyuan hy4 to \$838 for GPT-6-astra.

\section{Model-wise Model Trajectory Analysis}
\label{app:failure-domain-analysis}


\begin{figure}[!tb]
  \centering
  \includegraphics[width=\linewidth]{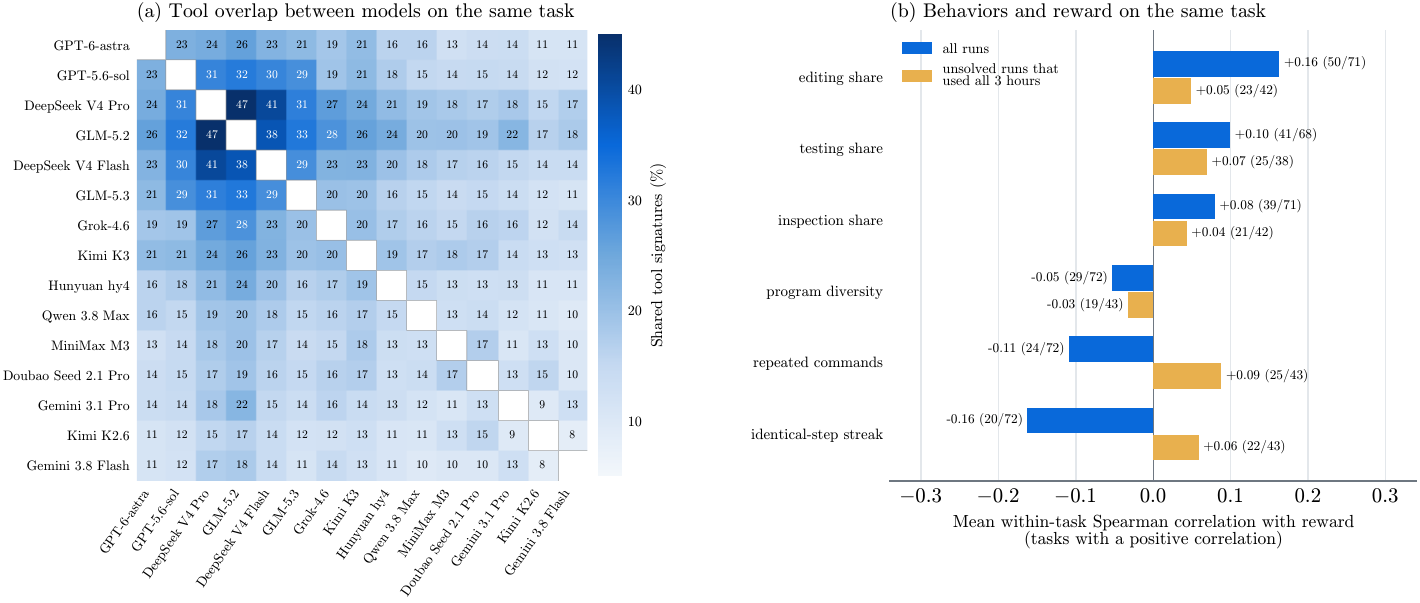}
  \caption{\textbf{Models use similar toolsets, but stronger runs edit more and repeat less.}
  (a)~The task constrains the toolset, but models still take different paths. Share of tool signatures (program plus targeted script or subcommand) that two models' runs on the same task have in common, averaged over tasks; models are ordered by similarity.
  (b)~Higher-reward runs tend to edit, test, and inspect more, while repeated-command loops correlate with lower reward; program diversity matters little. Spearman correlation of each command statistic with reward across models on the same task, averaged over tasks; the label counts tasks with a positive correlation. The two Gemini submissions, which ran under a different harness, are excluded.}
  \label{fig:lhtb-trajectory}
\end{figure}

We parse every shell command in the 1,135 runs that kept a full trajectory and compare how models work out the tasks rather than how well they do, where run is described by its mix of command types.
%

\paragraph{Different models use similar toolsets, but not the same commands.}
When two different models attempt the same task, their trajectories share only $3\%$ of exact commands and $18\%$ of tool signatures on average, even though their overall mixes of programs are much more similar (cosine similarity $0.63$). 
In other words, the task largely determines which tools are used, but the specific command sequence still depends strongly on the model.

\paragraph{Simple command statistics separate successful runs from unsuccessful ones, but not stronger failures from weaker failures.}
Across all runs, trajectories that spend more commands editing files tend to score higher, while trajectories with long repeated-command streaks tend to score lower (Figure~\ref{fig:lhtb-trajectory}b). 
These statistics mainly help distinguish runs that eventually solve the task from runs that do not. 
However, among the 800 runs that still fail after using up the full three-hour budget, they do not reliably indicate which runs made more progress than others.

\FloatBarrier

\section{Related Work}
\label{sec:related}

\paragraph{Long-horizon terminal tasks and trajectories.}
Long-horizon terminal benchmarks move beyond isolated code generation to tasks that require agents to inspect files, invoke tools, use software, and recover from errors over many steps. 
Benchmarks such as Terminal-Bench 2.1, Terminal Bench 4.0, Long-Horizon-Terminal-Bench, and LongCLI-Bench progressively raise the difficulty and horizon of these evaluations~\cite{merrill2026terminalbenchbenchmarkingagentshard,feng2026longcli}. 
As task horizons grow, reliable completion becomes harder~\cite{kwa2025measuring}, and performance depends not only on the model but also on the harness that provides tools, manages context, and handles execution feedback~\cite{yao2026harness,hu2026beyondmodel}. 
At the same time, longer and more realistic tasks are harder to support with high-quality reference solutions and agent trajectories, because collecting and validating them becomes increasingly expensive and time-consuming~\cite{li2026recursivesynthesislonghorizonterminal,chu2026terminalworldbenchmarkingagentsrealworld}. 
Progress in long-horizon evaluation therefore depends not only on constructing harder and more diverse tasks, but also on building the solution and trajectory resources needed to understand how agents succeed or fail on them~\cite{li2026finalscoressystematicevaluation,li2026scalingtrajectoriescomplextasks}.

\paragraph{Long-horizon task evaluation and training.}
For long horizon tasks, an unsuccessful final state may still contain completed sub-tasks and useful work.
Dense, verifiable process rewards credit intermediate achievements, while repeated checkpoints show whether progress continues, stalls, or is lost~\cite{li2026longhorizonterminalbenchtestinglimitsagents,li2026finalscoressystematicevaluation}. 
This distinguishes an agent that nearly finishes from one that makes little headway, and helps locate failures in planning, implementation, or feedback use~\cite{li2026finalscoressystematicevaluation}. 
Intermediate rewards also address sparse credit assignment during reinforcement learning by supplying feedback before a long rollout ends~\cite{lightman2023letsverify,yao2026earlyreward,yang2026t1terminalagentreinforcement}. 
Recent work on early reward prediction reports gains in coding-agent RL while reducing rollout cost~\cite{yao2026earlyreward}, showing the important of process rewards on long-horizon tasks.

\section{Conclusion}
\label{sec:conclusion}
\ours{} extends the original LHTB to 77 terminal tasks spanning 13 domains and is designed to evaluate long-horizon execution.
Across 15 frontier models evaluated with a shared \texttt{terminus-2} agent, the strongest model, GPT-6-astra, resolves only $48.1\%$ of tasks ($37/77$) at $R \geq 0.95$, while only 192 of 1,145 scored runs ($16.8\%$) pass overall.
At the same time, outcome-only evaluation misses substantial progress: many runs continue improving over long horizons, process reward often rises steadily even on runs that never fully pass, and a large fraction of total cost is spent in the final hour of unsolved runs.
These results suggest that the main challenge in long-horizon terminal tasks is not only local reasoning, but also sustaining productive work, preserving progress, and converting partial solutions into reliable completion over time.

\bibliographystyle{plainnat}
\bibliography{refs}

\clearpage
\appendix
\section{Task Inventory}
\label{app:task-list}

Table~\ref{tab:lhtb-task-list} lists all 77 benchmark tasks using the 13-domain taxonomy used throughout the paper. The first 46 rows preserve the original LHTB suite; the remaining 31 rows are the LHTB 2.0 additions, comprising 30 reconstruction tasks and one optimization task. Difficulty is derived from dense mean reward across model results in Figure~\ref{fig:lhtb-leaderboard}: tasks with mean reward at least $0.5$ are labeled \emph{Easy}, and the rest are labeled \emph{Hard}.

\small
\setlength{\LTleft}{0pt}
\setlength{\LTright}{0pt}
\setlength{\LTcapwidth}{\textwidth}
\setlength{\tabcolsep}{3pt}
\let\lhtborigunderscore\_
\def\_{\lhtborigunderscore\allowbreak}
\begin{longtable}{%
  >{\raggedright\arraybackslash}p{0.20\textwidth}%
  >{\raggedright\arraybackslash}p{0.18\textwidth}%
  >{\raggedright\arraybackslash}p{0.44\textwidth}%
  >{\raggedright\arraybackslash}p{0.10\textwidth}}
\caption{\textbf{Task list for \ours{}.} Each row shows a task ID, its paper-level category, a one-sentence description, and a coarse difficulty label derived from average task reward in Figure~\ref{fig:lhtb-leaderboard} (\emph{Easy} if mean reward $\geq 0.5$, otherwise \emph{Hard}).}
\label{tab:lhtb-task-list}\\
\toprule
\textbf{Task ID} & \textbf{Category} & \textbf{Description} & \textbf{Difficulty} \\
\midrule
\endfirsthead

\toprule
\textbf{Task ID} & \textbf{Category} & \textbf{Description} & \textbf{Difficulty} \\
\midrule
\endhead

\bottomrule
\endfoot

2048 & Interactive games & Play 2048 turn-by-turn via a terminal game server, merging tiles to reach the highest tile, scored by replaying recorded moves in a fresh seeded engine. & Hard \\
alp-paper-reproduction & Research reproduction \& ML & Reproduce a calibrated ALP floating-point compression experiment from the paper's method and sample columns. & Hard \\
apex-ib244-matter & APEX professional workflows & Advise on taking EdTech firm KSchool private across 38 reactive stages, rebuilding LBO/DCF/comps models for exact figures. & Easy \\
apex-investment-banking-matter & APEX professional workflows & Work a private-equity take-private of Planet Fitness through 19 reactive stages, rebuilding an LBO model and answering exact figures. & Hard \\
apex-law433-matter & APEX professional workflows & Act as outside counsel across four legal matters and 70 reactive stages over \textasciitilde{}600 documents, submitting deterministically-graded answers. & Easy \\
apex-management-consulting-matter & APEX professional workflows & Work two management-consulting client matters across 33 reactive stages, rebuilding business-case and cost-savings models for exact answers. & Hard \\
apex-openroad-ibex-signoff & Software \& reverse engineering & Close RTL-to-GDSII signoff on the ibex CPU with OpenROAD/sky130 across reactive stages until timing, DRC, LVS, area, and power pass. & Hard \\
audio-visual-event-alignment & Multimodal \& imaging analysis & Repair an audio-video sync audit that detects visual events, audio onsets, dropped frames, and per-clip synchronization offsets. & Hard \\
chess-mate & Interactive games & Play chess as White against a bundled superhuman Stockfish engine, scored on how long the agent survives before being checkmated. & Hard \\
climate-netcdf-extreme-event-audit & Earth, climate \& energy & Repair a climate audit pipeline computing gridded heatwave, precipitation, drought, anomaly, and index metrics over NetCDF datasets. & Hard \\
commit0-multilib-tdd & Software \& reverse engineering & Reimplement three whole Python libraries from stubs with no provided tests, writing your own suite, graded on hidden real test suites. & Hard \\
dicom-radiology-audit & Multimodal \& imaging analysis & Audit synthetic DICOM CT series geometry, HU conversion, windowing, masks, and lesion measurements. & Hard \\
document-table-layout-reconstruction & Multimodal \& imaging analysis & Repair a scanned-document pipeline that detects table grids, assigns noisy OCR tokens to cells, and emits CSV/JSON/Markdown tables. & Hard \\
duckdb-optimizer-closure & Systems, performance \& security & Iteratively patch DuckDB's query optimizer to speed up TPC-H while preserving correctness. & Hard \\
epa-swmm-stormwater-regression-audit & Earth, climate \& energy & Audit EPA SWMM stormwater hydraulic and hydrologic simulations against official solver outputs. & Hard \\
epidemic-inverse-control-audit & Scientific computing \& simulation & Fit and control a multi-city, age-structured SEIR-H-ICU epidemic model. & Hard \\
foldseek-paper-reproduction & Research reproduction \& ML & Reproduce a calibrated Foldseek-style structural protein search experiment from the Nature Biotechnology paper. & Hard \\
gdal-proj-raster-regression & Earth, climate \& energy & Reproduce GDAL/PROJ raster reprojection, coordinate transform, point sampling, and zonal-statistics regression checks. & Hard \\
generals-bot-arena & Software \& reverse engineering & Develop a bot for a Generals.io-style grid strategy game by iteratively playing matches and refining its strategy. & Easy \\
grammar-fuzz-coverage-hunt & Systems, performance \& security & Iteratively build a grammar-guided fuzzer to maximize code coverage of target parsers. & Easy \\
great-expectations-audit & Software \& reverse engineering & Build a Great Expectations-based multi-table data-quality and reconciliation audit pipeline. & Hard \\
langchain-version-migration & Software \& reverse engineering & Migrate a legacy LangChain RAG/router app across a breaking API change with hard-fail compatibility gates. & Hard \\
materials-phase-diagram-audit & Scientific computing \& simulation & Repair a materials phase-diagram audit reconstructing phase boundaries, invariant reactions, lever-rule fractions, and convex-hull stability. & Hard \\
matpower-opf-regression & Earth, climate \& energy & Reproduce MATPOWER/PYPOWER power-flow and optimal-power-flow regression checks with grid-constraint diagnostics. & Hard \\
microscopy-cell-count-qc-audit & Multimodal \& imaging analysis & Repair a microscopy QC audit that segments synthetic cell imagery, flags focus/artifact failures, and reports per-tile channel counts. & Hard \\
modflow6-groundwater-regression-audit & Earth, climate \& energy & Audit USGS MODFLOW 6 groundwater simulations against official solver outputs. & Hard \\
nbody-accel-iterative & Scientific computing \& simulation & Iteratively optimize an N-body gravitational simulation from O(N\textasciicircum{}2) to O(N log N) with correctness verification at multiple scales. & Easy \\
nrel-pysam-hybrid-renewables-audit & Earth, climate \& energy & Audit renewable generation and storage dispatch against NREL PySAM official PVWatts outputs. & Hard \\
opensees-seismic-structural-regression-audit & Scientific computing \& simulation & Audit seismic structural-dynamics results against official OpenSeesPy finite-element solver outputs. & Hard \\
poc-exploit-craft & Systems, performance \& security & Generate proof-of-concept inputs that trigger specific vulnerability classes using sanitizer feedback. & Easy \\
riscv-core-debug & Software \& reverse engineering & Localize and fix undisclosed injected bugs in an out-of-order RISC-V SystemVerilog CPU until simulation, compliance, and formal checks pass. & Easy \\
robotics-slam-benchmark-repair & Research reproduction \& ML & Repair a SLAM benchmark audit that composes SE(2) poses, evaluates loop closures, reconstructs landmark maps, and reports trajectory consistency. & Hard \\
rush\_hour\_campaign & Logic \& constraint puzzles & Solve a four-stage 6x6 Rush Hour campaign by hand, submitting legal move routes under length caps. & Hard \\
satellite-flood-change-detection-audit & Multimodal \& imaging analysis & Repair a satellite flood change-detection audit that compares before/after imagery, masks clouds, and reports per-tile flood extent. & Hard \\
scientific-figure-data-reconstruction & Multimodal \& imaging analysis & Repair a pipeline that reconstructs plotted series data from scientific figure images, captions, and partial CSV references. & Hard \\
snake\_maze\_campaign & Interactive games & Play a deterministic hard Snake variant, eating seeded apples and surviving growing obstacles as long as possible, scored by replaying the move log. & Hard \\
sokoban & Logic \& constraint puzzles & Solve a ramp of Sokoban puzzles turn-by-turn through a game server, advancing as far as possible, scored by replaying moves in a clean engine. & Hard \\
spice-ephemeris-regression & Scientific computing \& simulation & Reproduce NASA SPICE ephemeris, angle, and ground-station visibility regression checks from pinned kernels. & Hard \\
spot-scheduler-traces & Systems, performance \& security & Implement and refine a cloud spot-instance scheduling policy that minimizes cost while meeting hard deadlines. & Easy \\
su2-airfoil-regression & Scientific computing \& simulation & Reproduce a hardened SU2 aerodynamic regression matrix from official test-case references and solver history artifacts. & Hard \\
sudoku-recovery & Logic \& constraint puzzles & Recover seven progressively corrupted hard variant-Sudoku puzzles turn-by-turn, detecting subtle corrupt givens via global constraint reasoning. & Hard \\
super-mario & Interactive games & Play Super Mario Bros turn-by-turn through a game server, clearing as many levels as possible, scored by replaying frame inputs in a fresh emulator. & Hard \\
tabular-data-feature-covshift & Research reproduction \& ML & Recover a sparse high-dimensional signal from low-dimensional observations under covariate shift, with a leakage-audited honesty gate. & Hard \\
unison-paper-reproduction & Research reproduction \& ML & Reproduce a calibrated UNISON fat-tree network simulation experiment from the paper's method and settings. & Hard \\
unknown-config-semantics & Software \& reverse engineering & Reverse-engineer an undocumented config format from a noisy spec and repair its normalization engine to pass hidden probes across five stages. & Hard \\
vector-db-iterative-build & Systems, performance \& security & Build an approximate nearest-neighbor vector search service from scratch, iteratively optimizing recall and throughput. & Hard \\
acoustic\_beamforming\_peak\_\_01 & Physical simulation, signals \& control & Reconstruct a room-acoustics beamforming pipeline so its peak-direction estimates match a reference simulator. & Hard \\
ajv\_dynamic\_ref\_tree\_\_05 & Software \& reverse engineering & Reconstruct AJV JSON-schema validation behaviour over dynamic \$ref trees from observed accept/reject traces. & Hard \\
astro\_mosaic\_reprojection\_\_04 & Multimodal \& imaging analysis & Reconstruct an astronomical FITS mosaic reprojection and photometry pipeline from reference outputs. & Hard \\
chem\_reaction\_smarts\_application\_\_02 & Life sciences, chemistry \& materials & Reconstruct SMARTS reaction-template application so product sets match a reference cheminformatics engine. & Hard \\
circuit\_buck\_converter\_ripple\_\_01 & Physical simulation, signals \& control & Reconstruct a buck-converter circuit simulation reproducing reference ripple and transient measurements. & Hard \\
climate\_regrid\_conservation\_check\_\_06 & Earth, climate \& energy & Reconstruct a NetCDF climate regridding pipeline that preserves the reference conservation diagnostics. & Hard \\
duckdb\_recursive\_hierarchy\_report\_\_03 & Document, OCR \& data engineering & Reconstruct a DuckDB recursive-hierarchy report matching reference aggregation output. & Hard \\
fea\_plate\_hole\_stress\_concentration\_\_01 & Physical simulation, signals \& control & Reconstruct a finite-element plate-with-hole analysis reproducing reference stress-concentration factors. & Hard \\
font\_harfbuzz\_shaping\_cluster\_\_04 & Document, OCR \& data engineering & Reconstruct HarfBuzz text-shaping cluster and glyph-advance behaviour from observed shaping output. & Hard \\
gdal\_antimeridian\_clip\_\_01 & Geospatial \& cartography & Reconstruct GDAL antimeridian-crossing clip geometry matching reference vector output. & Hard \\
gdal\_vrt\_mosaic\_transform\_\_03 & Geospatial \& cartography & Reconstruct a GDAL VRT mosaic and coordinate transform reproducing reference raster statistics. & Hard \\
genetic-convergence-testing & Scientific computing \& simulation & Optimize a stochastic convergence-testing workflow against hidden statistical and runtime criteria. & Hard \\
geology\_multiwell\_correlation\_panel\_\_04 & Earth, climate \& energy & Reconstruct a multi-well log correlation panel matching reference stratigraphic picks. & Hard \\
ghostscript\_pdfa\_preflight\_repair\_\_06 & Document, OCR \& data engineering & Reconstruct a Ghostscript PDF/A preflight-and-repair pipeline matching reference compliance reports. & Hard \\
hydro\_hydrograph\_response\_\_06 & Earth, climate \& energy & Reconstruct a raster hydrology model reproducing reference hydrograph response curves. & Easy \\
lidar\_icp\_registration\_summary\_\_06 & Geospatial \& cartography & Reconstruct LiDAR ICP point-cloud registration reproducing reference alignment summaries. & Hard \\
map\_label\_priority\_collision\_\_04 & Geospatial \& cartography & Reconstruct cartographic label placement and collision priority matching reference rendered labels. & Hard \\
map\_road\_class\_generalization\_\_05 & Geospatial \& cartography & Reconstruct road-class cartographic generalization matching reference simplified geometry. & Hard \\
materials\_xrd\_peak\_shift\_\_04 & Life sciences, chemistry \& materials & Reconstruct X-ray diffraction peak-shift analysis reproducing reference lattice measurements. & Hard \\
medical\_slice\_series\_reassembly\_\_03 & Multimodal \& imaging analysis & Reconstruct medical slice-series reassembly and geometry recovery matching reference volumes. & Hard \\
nginx\_location\_regex\_tryfiles\_precedence\_\_06 & Systems, performance \& security & Reconstruct nginx location/regex/try\_files routing precedence from observed request traces. & Hard \\
opa\_rbac\_namespace\_boundary\_\_05 & Systems, performance \& security & Reconstruct OPA/Rego RBAC namespace-boundary policy decisions from observed allow/deny traces. & Hard \\
openssl\_pkits\_name\_constraints\_\_05 & Systems, performance \& security & Reconstruct OpenSSL PKITS name-constraint certificate validation from reference verdicts. & Easy \\
pcap\_http2\_tls\_fingerprint\_\_04 & Systems, performance \& security & Reconstruct HTTP/2 and TLS fingerprint extraction from packet captures matching reference fingerprints. & Hard \\
pdb\_multimodel\_rmsd\_\_02 & Life sciences, chemistry \& materials & Reconstruct multi-model PDB structural RMSD computation matching reference superposition values. & Hard \\
proteomics\_retention\_time\_alignment\_\_02 & Life sciences, chemistry \& materials & Reconstruct mass-spectrometry retention-time alignment matching reference peptide mappings. & Hard \\
robotics\_pick\_reach\_controller\_\_01 & Physical simulation, signals \& control & Reconstruct a robotic pick-and-reach controller reproducing reference trajectories and success criteria. & Hard \\
samtools\_mpileup\_consensus\_\_05 & Life sciences, chemistry \& materials & Reconstruct a samtools mpileup consensus-calling pipeline matching reference variant output. & Hard \\
seismo\_cross\_correlation\_lag\_\_01 & Physical simulation, signals \& control & Reconstruct seismic waveform cross-correlation lag estimation matching reference arrival offsets. & Easy \\
sqlite\_wal\_recovery\_audit\_\_02 & Document, OCR \& data engineering & Reconstruct a SQLite WAL recovery audit reproducing reference post-crash database state. & Hard \\
tesseract\_table\_image\_ocr\_\_06 & Document, OCR \& data engineering & Reconstruct a Tesseract table-image OCR pipeline matching reference extracted tables. & Hard \\
\end{longtable}
\let\_\lhtborigunderscore
\normalsize

\end{document}